\documentclass[10pt,twocolumn,letterpaper]{article}

\usepackage{wacv}              

\newcommand{\R}{\mathbb{R}}

\usepackage{pifont}
\usepackage{lipsum}
\usepackage{multicol}
\usepackage{booktabs}
\usepackage[table]{xcolor}

\definecolor{wacvblue}{rgb}{0.21,0.49,0.74}
\usepackage[pagebackref,breaklinks,colorlinks,allcolors=wacvblue]{hyperref}

\def\wacvPaperID{1348} 
\def\confName{WACV}
\def\confYear{2027}

\title{Towards Zero-Shot Transfer Across Embodiments For Driving VLAs}

\author{Caio Azevedo$^{1, 2}$ \hspace{0.3cm}  Stefano Sabatini$^{2}$ \hspace{0.3cm} Sascha Hornauer$^{1}$ \hspace{0.3cm}  Fabien Moutarde$^{1}$  \smallskip \\
$^{1}$École des Mines de Paris, France\\
$^{2}$Stellantis, France \\
{\tt\small caio.azevedo@minesparis.psl.eu}
}

\begin{document}
\maketitle

\begin{abstract}
    Vision-Language-Action models (VLAs) have shown strong potential in autonomous driving by leveraging multimodal pretraining for instruction following, visual reasoning, and scene-level generalization.
    In robotic manipulation, scaling VLA fine-tuning across multiple robot setups--especially when unifying representations across embodiments--has been shown to improve in-dataset performance and cross-embodiment generalization; in autonomous driving, however, VLAs remain largely trained on individual datasets and are rarely evaluated for zero-shot transfer to unseen datasets and camera rigs; furthermore naively adding more datasets to the training data does not necessarily lead to better performance within seen embodiments.
    To address these problems, we study multi-dataset training for the driving task and BEV-Forcing, an auxiliary objective that transfers ground-plane object-layout information from a specialized Bird’s-Eye-View model into the VLA backbone.
    By encouraging the model to represent object position through a shared BEV spatial interface, we show that an auxiliary task such as BEV-Forcing can improve both in-distribution and out-of-distribution performance when training on a small number of camera rigs. As the number of training embodiments increases, however, the benefits of the auxiliary task are reduced; we present this as evidence that new techniques in the literature may see their benefits diminish when simply scaling up training diversity, which motivates presenting results taking into account data scaling.
    Code available in \url{github.com/caiocj1/ad-vla}.

\end{abstract}

\begin{figure*}[t]
        \centering
        \includegraphics[width=0.95\linewidth]{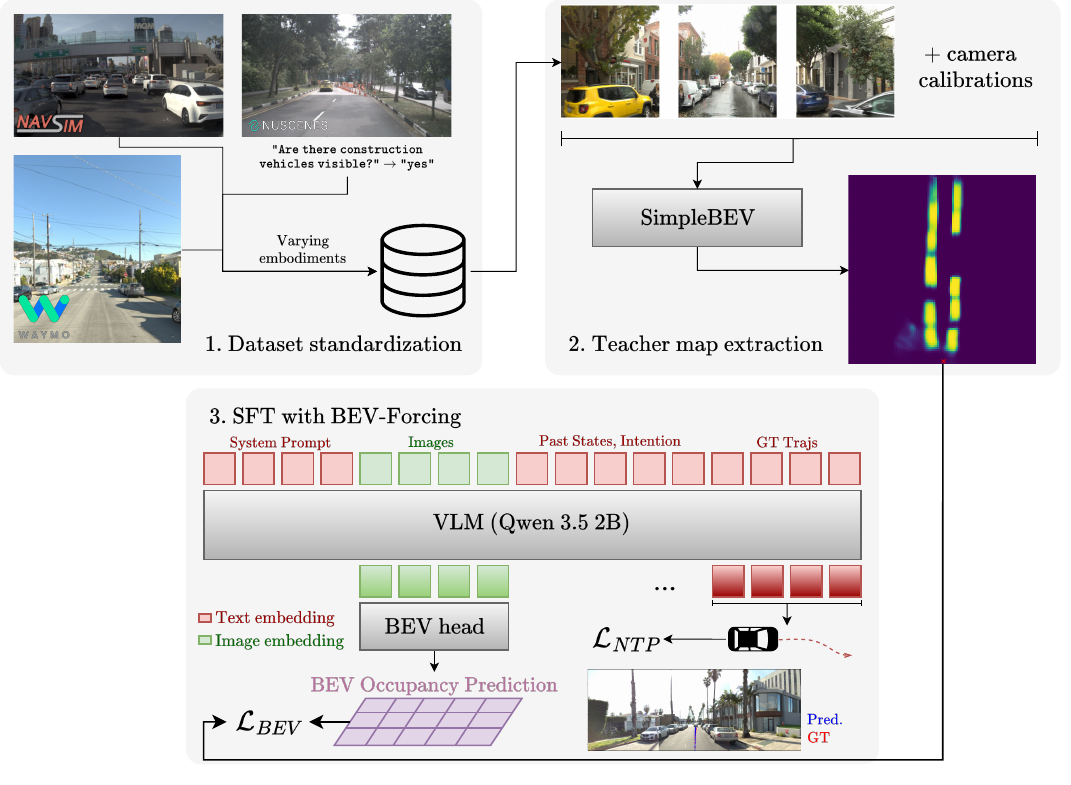}
            
        \caption{Illustration of BEV-Forcing, performing supervised fine-tuning (SFT) on a planning sample (analogous for VQA samples). After unifying various end-to-end datasets into a common format, we extract teacher occupancy maps from a specialized BEV model, SimpleBEV \cite{harley2023simple} in our case. These are used as the ground-truth for BEV-Forcing, which supervises last-layer image hidden states with extracted teacher maps, transferring ground plane object position awareness into the backbone.}
        \label{fig:schematics}
    \end{figure*}

\section{Introduction}

    Recent progress in the capabilities of vision-language models (VLMs) has opened up a promising new direction of research for end-to-end autonomous driving, namely, how to properly adapt VLM backbones into good driving vision-language-action models (VLAs). This problem has piqued the interest of the research community for several reasons: first, the large-scale multi-modal pretraining of the backbone provides models with semantic priors that can help in correctly interpreting the driving environment based on image content alone, even in rare or unseen long-tail scenes; they also are able to provide reasoning traces that can ground actions on the reality of the traffic scene, which in turn provides explainability and can in principle improve performance in hard scenarios; finally, they allow for human-machine interaction through their acquired understanding of natural language \cite{jiang2025survey}.

    In practice, the incorporation of pre-trained language backbones into embodied tasks has been shown, in both robotics \cite{zitkovich2023rt, kim2025openvla, black2025pi} and driving \cite{pan2024vlp, hwang2025emma}, to lead to both better in-dataset performance and better \textit{semantic transfer} than previous approaches, i.e. retaining performance over visual or scene-content distribution shifts, such as when novel objects appear in the scene or when a model is trained or fine-tuned in a certain location and evaluated on another using the same sensor rig. Additionally, modern fine-tuning techniques further improve performance on out-of-distribution tasks on the same embodiment, e.g. by representing actions \cite{hancock2026actions} or by adaptively fine-tuning parameters \cite{huang2026maps} in such a way as to avoid catastrophic forgetting. 
    
    However, when VLAs are adapted for robotic tasks such as driving, a problem arises: as the model is fine-tuned to output future actions, it might acquire the ability to spatially interpret the information from the sensor rig of the particular dataset it is being fine-tuned on, but catastrophically fail when asked to output spatially grounded trajectories on another dataset collected with a different camera setup.
    In other words, VLAs lack \textit{geometric transfer}: although the backbone may semantically understand the contents of the image from the new dataset, 3D spatial understanding does not necessarily transfer, since it is learned implicitly from a low number of sensor rigs and from much fewer examples than are present in the pretraining of the backbone.

    In robotics, it has been shown that pre-training on a wide variety of datasets from different embodiments alleviates this problem, while overall still demanding fine-tuning on the target setup \cite{zitkovich2023rt, kim2025openvla, doshi2025scaling}. In autonomous driving, efforts to unify the available open end-to-end driving datasets are still very recent \cite{chi2026impromptu, dauner2026123d}. They also show the potential of training on multiple datasets; however, most works remain trained on a single driving dataset, and if we take an ``embodiment" in autonomous driving to mean the characteristics of a particular sensor rig mounted on the ego-vehicle, we observe that not all training data compositions in terms of embodiments lead to improvements. Finally, further fine-tuning is still required for good performance when transferring to unseen camera rigs. The question of what makes a model good at cross-embodiment zero-shot transfer, in particular for autonomous driving, remains underexplored; this is an important problem since better zero-shot transfer allows for more effective adaptation into new embodiments and therefore facilitates prototype development and iteration.

    Given this context, we wish to \textit{investigate the behavior and performance of driving VLAs, both within seen camera rigs and crucially at zero-shot transfer, when we scale the number of training embodiments}. In particular, we hypothesize that improved zero-shot transfer may arise if the model is able to learn a unified spatial representation from multi-view camera images coming from varying rigs. To test this, we also \textit{study combining scaling training embodiments with an auxiliary objective related to spatial awareness}.

    In short, we propose the following contributions:
    
    \begin{itemize}
        \item BEV-Forcing, an auxiliary task inspired by Spatial Forcing \cite{li2025spatial}, which can improve both in and out of dataset performance when training on a limited number of camera rigs. This incurs no extra cost at inference time since the BEV auxiliary prediction head can simply be removed in a production environment, and is scalable since it does not require annotated maps.
        \item We provide our observations when scaling up the training of driving VLAs with multiple public datasets, including planning and visual-question-answering (VQA) tasks, evaluating performance in both seen embodiments and at zero-shot transfer with and without BEV-Forcing and several different training configurations. We show how applying BEV-Forcing when training on few embodiments can lead to gains comparable to those obtained by adding extra training embodiments.
        \item Finally, we provide a series of ablations to better understand the important elements of the auxiliary task and how it interacts with other techniques aimed at increasing robustness such as performing image augmentations or passing calibration parameters as inputs.
    \end{itemize}

\section{Related Work}

    \textbf{Spatial awareness for VLAs.} A VLA needs to acquire the knowledge of how to properly navigate a 3D environment, since this knowledge is not sufficiently acquired during the large-scale pretraining of its backbone. For this reason, a variety of approaches try to teach 3D awareness to VLAs. In many works in autonomous driving, this is learned implicitly \cite{rowe2025poutine, zhou2026autovla, wang2025alpamayo}. This is enough for strong in-dataset performance results, though edge cases still provide challenges which motivate the development of explicit spatial awareness learning; additionally, performance when transferring to a new embodiment is usually not evaluated.
    
    Passing LiDAR-to-camera aligned projection images as an extra input to the backbone has been shown to improve question-answering accuracy on driving scenes \cite{han2026d3vl}. On planning, other approaches also incorporate extra modalities into the model: (i) with 3D data supervision as an auxiliary objective, either of geometry modules as extra parameters \cite{fu2025orion, yao2026vlga} or of the backbone itself, but this significantly increases data collection requirements; (ii) by treating the new modalities as extra inputs that contain explicit spatial information that the model learns to use after multi-stage training \cite{zhou2026opendrivevla, winter2025bevdriver}, but this does not take full advantage of the strong vision-language alignment of current natively multi-modal models \cite{qwen3.5}; (iii) by taking advantage of target features that can instill spatial awareness into the model, whether from 3D foundation models such as VGGT \cite{wang2025vggt} and used as direct supervision targets \cite{li2025spatial, luo2026last}, or from trained dynamics-aware future image predictors \cite{shang2026dynvla}, or still by conditioning on VGGT features with cross-attention  \cite{wang2026vggdrive}.
    
    In our work, we decide to instill spatial awareness using camera inputs only and no 3D data label requirement: first, to fully leverage the vision-language alignment of recent VLM backbones without extra training stages; second, due to the scarceness and cost of collection of 3D data such as LiDAR: important datasets such as WOD-E2E \cite{xu2026wod} only provide camera images. Our approach aligns most with category (iii), and in particular with Spatial Forcing \cite{li2025spatial}. We decide however to use as targets for representation supervision occupancy maps output by specialized BEV models, as we hypothesize that these may contain a more uniform and more task aligned spatial interface across embodiments compared to 3D depth-map reconstruction related features. This decision is justified in Section \ref{sec:ablations}.

    \textbf{VLA generalization.} Many works, especially in robotic manipulation, have demonstrated and further deepened the capability of VLAs for semantic zero-shot transfer, i.e. evaluating a model on an embodiment it has seen during training but in a new environment or task.
    In particular, large-scale pretraining on a variety of embodiments is of fundamental importance \cite{o2024open, kim2025openvla, black2025pi, zheng2026x}.
    Recent techniques such as adapting the constraint on parameter updates depending on model depth \cite{huang2026maps}, casting robotic tasks as pure natural language \cite{hancock2026actions}, among many others \cite{li2026vla} help retain the knowledge of the backbone's pretraining leading to better semantic transfer.
    
    With respect to geometric or cross-embodiment zero-shot transfer, there is a relative lack of literature in autonomous driving, though recent efforts are trying to cover this gap \cite{dauner2026123d, chi2026impromptu, wagner2026longtail}. Previous works in autonomous driving have shown that world modeling through video generation aligned with actions unlocks new scaling laws and benefits zero-shot transfer across datasets \cite{li2026drivevla, liu2026driveva}, but this leads to high computational requirements both at training and at inference time. In robotic manipulation, the emphasis is either on learning embodiment specific adapters \cite{zheng2026x} or on the necessity of unifying observations and action representations across many embodiments and tasks as a means of transferability \cite{yuan2026qwen, qu2025spatialvla}.

    Autonomous driving presents its own set of both challenges and advantages compared to robotic manipulation: while there are fewer large-scale public datasets to provide camera setup variability, having a unified task with similar outputs across embodiments (planned trajectories) should in principle facilitate scaling up training to multiple datasets with positive transfer, independently of the particular action representation being used by the backbone. For these reasons, in our work we settle on a simple textual action representation (which is nevertheless argued for in \cite{hancock2026actions}) and focus on investigating a representation learning technique that can improve zero-shot transfer when training on a limited number of camera rigs and with no extra inference cost, as well as investigating how this technique behaves as we train on more embodiments.

\section{Methodology}

    In this section we explain in detail the architecture and design choices we made for instilling spatial awareness with BEV-Forcing according to Fig. \ref{fig:schematics}, as well as how we perform multi-dataset training.
    
    \subsection{Backbone and Action Representation}

        We start from a pretrained open-source vision-language model, Qwen 3.5 2B \cite{qwen3.5}. In our work we decide to focus on good spatial representation learning as a means of transferability. Because of this, for action representation we simply use text outputs representing trajectory waypoints at 1 Hz as pairs of 2D coordinates, following \cite{rowe2025poutine}. These are then parsed and resampled with cubic splines towards the target sampling rate. Consistent with \cite{hancock2026actions}, since the use of text for action representation brings the task closer to the original pretraining of the backbone, we find that applying LoRA \cite{hu2022lora} to avoid catastrophic forgetting leads to better performance than full fine-tuning of the backbone.

    \begin{figure}[t]
        \centering
        \includegraphics[width=\linewidth]{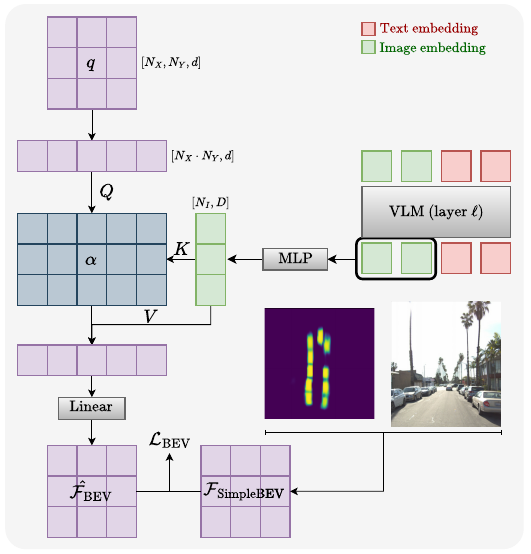}
            
        \caption{Illustration of the BEV head. Randomly initialized queries attend to image embeddings to output occupancy maps of vehicles in the scene. The low capacity of the BEV head forces image features that are used for trajectory planning to contain information related to surrounding objects' position in space.}
        \label{fig:bev_head}
    \end{figure}

    \subsection{BEV-Forcing}

        Similar to \cite{li2025spatial}, which supervises image embeddings with features from a specialized 3D model for more precise robotic manipulation, we aim to instill in the image embeddings within the model an awareness of the spatial representation of the scene, in a way that does not require extra computational costs at inference time. For this, we propose BEV-Forcing as described in Fig. \ref{fig:schematics}: we add a BEV head as shown in Fig. \ref{fig:bev_head}, that first initializes queries $q \in \R^{(N_X \cdot N_Y) \times d}$, with $N_X \times N_Y$ being the resolution of the spatial grid and $d$ the hidden dimension of the BEV head.
        We extract image embeddings $f_I^{(\ell)} \in \R^{N_I \times D}$ at a target layer $\ell$ of the backbone, then project these with an MLP to the query dimension $d$, $N_I$ being the number of image tokens and $D$ the dimension size of the backbone.
        Then the queries cross-attend to the projected image embeddings. Finally, the post-attention features $\hat{b}$ are projected to occupancy logits at each position through a simple linear layer:
        \begin{equation}
            \tilde{f_I} = \text{MLP}(f_I^{(\ell)}),
            \label{eq:embed_img_fts}
        \end{equation}
        \begin{equation}
            \hat{b} = \text{Attn}(W_Q \cdot q, W_K \cdot  \tilde{f_I}, W_V \cdot \tilde{f_I}).
            \label{eq:post_attn_output}
        \end{equation}
        \begin{equation}
            \hat{\mathcal{F}}_\text{BEV} = \text{Linear}(\hat{b}).
            \label{eq:output_bev_attn}
        \end{equation}
        By limiting the capacity of the BEV auxiliary head that is added on to the backbone (the architecture and parameters in Eq. \ref{eq:embed_img_fts} to \ref{eq:output_bev_attn}), by for example simply using a cross-attention layer instead of a full transformer block, we ensure that the representation of the image tokens $f_I^{(\ell)}$ acquires as much information about the spatial disposition of objects in the scene as possible, which then can help in planning performance as will be shown in Section \ref{sec:exps}. During inference, the BEV head can simply be removed, leading to no extra computational cost.
        
    \subsection{Training}

        The predicted $\hat{\mathcal{F}}_\text{BEV}$ grid is then supervised with a binary cross entropy loss $\mathcal{L}_{\text{BEV}}$ with targets coming from the occupancy map ${\mathcal{F}}_\text{SimpleBEV}$ that is output by a specialized BEV model, SimpleBEV in our case \cite{harley2023simple}. SimpleBEV's architecture allows it to output reasonably high-quality occupancy maps even for datasets it was not trained on, and using a BEV expert model gives us a training signal in important datasets which do not have ground-truth maps such as WOD-E2E \cite{xu2026wod}. For loss calculation, positive grid positions have higher weight $w_+$ to compensate data imbalance.
        
        $\mathcal{L}_\text{BEV}$ is then weighted by a factor $\lambda$ and added to standard next-token prediction (NTP) loss during supervised fine-tuning, where the target sequence is simply the ground-truth future trajectory in text tokens:
        \begin{equation}
            \mathcal{L} = \mathcal{L}_{\text{NTP}} + \lambda \cdot \mathcal{L}_{\text{BEV}}
        \end{equation}

        To make sure that the model does not lose its linguistic capabilities and semantic knowledge, we also do runs with visual-question-answering (VQA) co-training. For VQA samples, the loss is obtained in the same manner: we calculate the next-token prediction loss using the expected answer for each question, while the target layer's image hidden states are passed to the BEV head to produce $\mathcal{L}_{\text{BEV}}$.

    \subsection{Dataset Standardization}

        We load samples from each dataset in a unified contract that easily allows combining multiple datasets. In particular, for training on planning samples we require past trajectories, current timestep images from the three camera images and high-level ego-vehicle intent as inputs, as well as the target future trajectory for supervised fine-tuning; for VQA samples, we pass as inputs all surrounding camera images at the current timestep and supervise on the single-word answer according to the format of nuScenes-QA \cite{qian2024nuscenes}. For both tasks, a natural language prompt (available in Supplementary Material) specifies the task and output format. Dataset interfaces are available in the code repository.

\section{Experiments}
\label{sec:exps}

    To recapitulate, we wish to study the behavior of a driving VLA as the number of training embodiments scales, on seen embodiments but more importantly at zero-shot transfer to a new embodiment. Additionally, we investigate the effect of applying BEV-Forcing across training configurations. Here we present and discuss our experiments.

    \subsection{Experimental Setup}

        \textbf{Datasets and Metrics.} For our experiments with planning samples we use WOD-E2E \cite{xu2026wod}, NAVSIM \cite{dauner2024navsim, cao2025pseudo}, nuScenes \cite{caesar2020nuscenes}, and a subset of six thousand clips of Physical AI \cite{nvidia2025physicalaiav} due to computational constraints. Each dataset provides camera images at the current timestep, past ego-vehicle states, and for some of the datasets the high-level intent of the ego-vehicle. For VQA we use nuScenes-QA \cite{qian2024nuscenes}.
        
        Our goal with using Physical AI specifically is for zero-shot evaluation and comparison with a model trained directly on Physical AI. For this, we randomly sample five thousand clips for training and one thousand for validation, each clip consisting of 20 seconds of video and other sensor data as well as the features mentioned above. Selected clips are available in code repository for reproducibility. The validation set is used as the target for zero-shot evaluation, while we use the five thousand clip training subset only in Fig. \ref{fig:phai1k-val-zeroshot}, i.e. so we can compare our models in zero-shot transfer with ones trained on Physical AI; there, 5k $\times$ 9 means training on 9 timesteps extracted at regular intervals from each of the five thousand clips, and analogously for 21. We also evaluate zero-shot on the test set KIT LongTail Dataset \cite{wagner2026longtail}, which was released for a limited time specifically for zero/few shot evaluation with no training set available.
        
        With regards to metrics, we use average and final displacement error (ADE/FDE), as well as the rater feedback score (RFS) on WOD-E2E \cite{xu2026wod} and the multi-maneuver score (MMS) on the KITScenes LongTail dataset \cite{wagner2026longtail}. ADE/FDE simply measure the average and final distance between prediction and recorded ground-truth planned trajectory respectively (lower is better). Both RFS and MMS are calculated based on whether predictions match with human annotated trajectories with assigned scores that take into account the trajectory's quality. For both RFS and MMS, higher is better.
        
        \textbf{Implementation Details.} We use LoRA with rank 16, a learning rate of $10^{-4}$ with cosine decay, and train for one epoch on all experiments, using global batch size 64 across 4 A100 GPUs. For each planning dataset, we pass the three front cameras as an input to the model for the planning task; for VQA we use all camera images, always downsizing to lower resolutions to reduce VRAM requirements while keeping the aspect ratio close to the original images; every image transformation is accompanied by its corresponding camera calibration parameters transformation. To extract teacher occupancy maps, we use higher resolution images to maximize the performance of SimpleBEV. Training runs all use 16 past time step trajectories at 4 Hz. We take $\ell$ as the last layer of the VLM backbone. More architectural details are present in the Supplementary Material.
        

            

    \subsection{Experimental Results}

         \begin{figure}[t]
                \hspace*{-0.7cm}
                \includegraphics[width=1.1\linewidth]{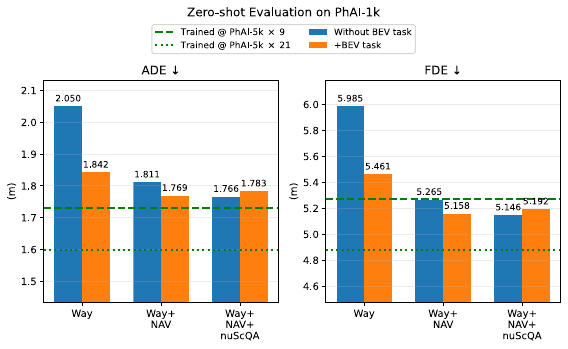}
            
                \caption{Performance on PhAI-1k validation set after training on different dataset combinations, with and without BEV task. Training on WOD-E2E only, the BEV task gives large improvements on zero-shot transfer. Introducing NAVSIM and subsequently nuScenesQA into training data also substantially helps zero-shot transfer, with benefits overlapping with those of BEV-Forcing as training variability increases.}
                \label{fig:phai1k-val-zeroshot}
            \end{figure}

\begin{figure}[t]
                \includegraphics[width=1\linewidth]{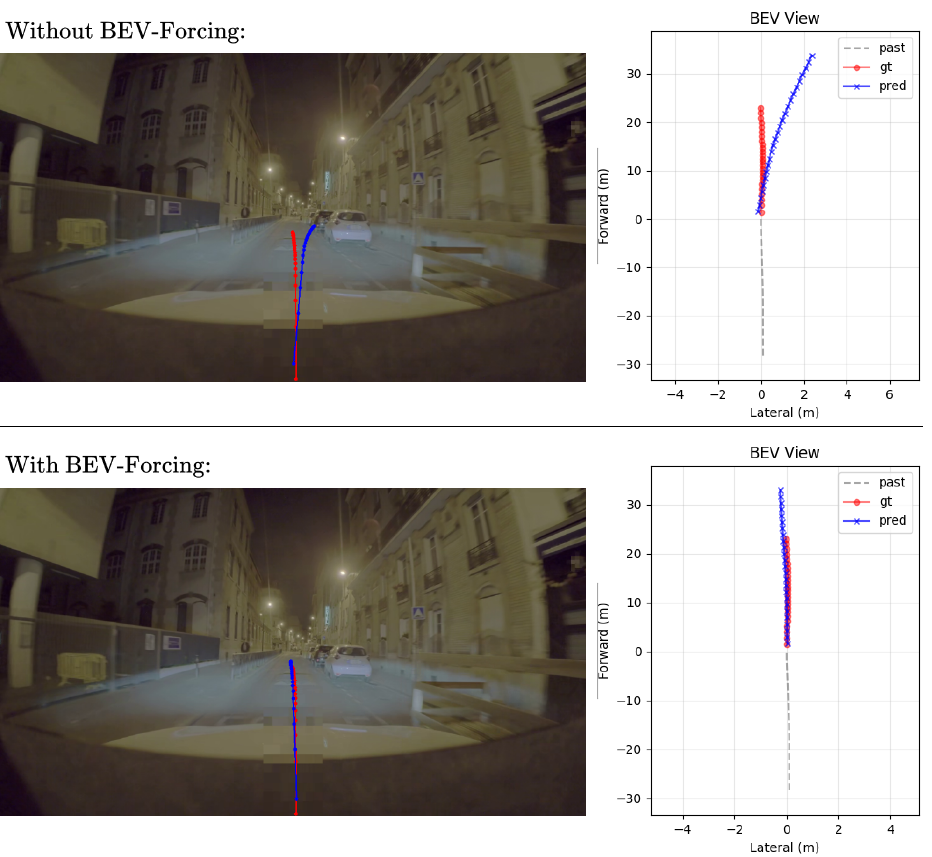}
            
                \caption{Qualitative case on a validation sample of Physical AI, of our model trained only on WOD-E2E, without the BEV task above and with it below. BEV-Forcing can adequately provide spatial information that allows the model to perform significantly better at zero-shot transfer.}
                \label{fig:phyai_quali}
            \end{figure}

        \textbf{BEV-Forcing helps zero-shot transfer when training on few embodiments.}
            First, we progressively increase the number of embodiments across separate training runs, from WOD-E2E only to then including NAVSIM and finally nuScenes-QA (we discuss the importance of including the VQA task below); we then compare for each training set of camera rigs, the performance with and without the BEV task. We evaluate zero-shot transfer planning performance using a thousand validation clips of Physical AI and on KITScenes LongTail. We observe that the BEV task can serve as a geometry-aware regularizer: it is able to provide large improvements when training on a limited number of camera rigs, with improvements diminishing as training-rig diversity increases. For instance, as seen in Fig. \ref{fig:phai1k-val-zeroshot}, applying the BEV loss while training on only WOD-E2E provides a 10.1\% decrease in ADE, an improvement which is seen in qualitative cases such as Fig. \ref{fig:phyai_quali}. Performance however comes closer to the baseline as we add datasets, suggesting that training diversity and BEV supervision may provide partially overlapping benefits.

            The same phenomenon is seen on the validation set of KITScenes LongTail, using an internal reproduction of the MMS  as described in its paper \cite{wagner2026longtail}. Training on WOD-E2E only, we observe an MMS of only 3.90 when not using the BEV task, which improves to 4.64 when adding it. However, as the number of training embodiments increases, data variability overshadows the effect of the BEV task: when training on WOD-E2E + NAVSIM + nuScenesQA, not using the BEV task gives an MMS of 5.13 while using it gives 4.84. For comparison with other methods, we evaluate our models on the test set of KITScenes. As seen in Table \ref{tab:kit-leaderboard}, our model trained only on WOD-E2E + NAVSIM, a relatively limited amount of variability, achieved an MMS of 5.15 on the test set when using the BEV task, an improvement over the baseline without the BEV task of 4.62 MMS. Our models outperform previous frontier models such as Gemini 3 Pro at zero-shot transfer, showing the need for specializing on domain data for properly navigating the environment; they also outperform solutions based on strong models such as Alpamayo 1.5 \cite{wang2025alpamayo}, which achieved 4.31 MMS. These results indicate that once again, BEV-Forcing offers useful signal for zero-shot transfer when training on a limited number of camera rigs.

            \begin{table}
            \centering
            \begin{tabular}{l|cc}
                \toprule 
                Method & MMS & L2 (mean)  \\
                \midrule
                \midrule
                UniAD \cite{hu2023planning} & 3.24 & 10.90 \\
                DMAD \cite{shen2025divide} & 3.51 & 10.04 \\
                Alpamayo 1.5 \cite{wang2025alpamayo}-based$^\dagger$ & 4.31 & 3.17 \\
                VLA \& Kinematics-based$^\dagger$ & 4.31 & 3.57  \\
                Gemini 3 Pro & 4.61 & 2.99 \\
                \midrule
                Ours [W+N] & {4.62} & {2.63}  \\
                Ours [W+N] (+BEV task) & \textbf{5.15} & \textbf{2.48}  \\
                \bottomrule
            \end{tabular}
            \caption{Performance of various methods on zero/few-shot evaluation at the test split of the KITScenes LongTail Dataset \cite{wagner2026longtail}. ``[W+N]" means our models are trained on WOD-E2E + NAVSIM. $^\dagger$Descriptions inferred from test set entry titles.}
            \label{tab:kit-leaderboard}
            \end{table}

        \textbf{BEV-Forcing helps in-dataset performance when training on a single embodiment.}
            Any auxiliary task aimed at improving zero-shot transfer must also not considerably degrade in-dataset performance. Due to constraints on submissions to the larger test set, we first evaluate all training configurations on the RFS-labeled validation split of WOD-E2E. Across these runs, BEV-Forcing consistently maintains or improves planning performance relative to the corresponding baseline. For example, when training on WOD-E2E only, FDE decreases from 5.946 to 5.831 by applying the BEV task, while RFS slightly increases from 7.961 to 7.971. Similar improvements are observed for the more diverse training configurations, with the best validation result using the BEV task reaching 5.687 FDE and 8.119 RFS, compared to 5.851 and 8.063 respectively for the baseline. These results indicate that the auxiliary spatial supervision does not introduce an apparent in-dataset performance trade-off on the validation split. We next evaluate selected configurations on the WOD-E2E test set to assess whether this behavior persists under the more reliable held-out evaluation.

            For this, we submit the models trained on WOD-E2E and on WOD-E2E + NAVSIM + nuScenes-QA.
            Results are in Table \ref{tab:wod-leaderboard}. First, notice that our overall performance is roughly consistent with the performance of the SFT-only Poutine model, which is expected due to the similar architectures. Next, notice that when training on a single embodiment, i.e. WOD-E2E, the BEV task is able to improve performance, while its effectiveness decreases when we add more diverse camera rigs. With both zero-shot and in-dataset results, an important take away is that techniques aiming at improving end-to-end planning can be useful on single-dataset training (the standard procedure across many papers), while in principle being susceptible to have their effectiveness decreased when training on more embodiments. We stress therefore the importance of evaluating techniques trained with multiple camera rig setups, including at zero-shot.

            \begin{table}
            \centering
            \begin{tabular}{l|cc}
                \toprule 
                Method & RFS & ADE  \\
                \midrule
                \midrule
                IRL-VLA \cite{jiang2025irl} & 7.890 & 2.823 \\
                Poutine (SFT-only) \cite{rowe2025poutine} & 7.909 & 2.941 \\
                Poutine \cite{rowe2025poutine} & 7.986 & 2.742 \\
                NTR \cite{li2026ntr} & 8.046 & 2.638 \\
                DriveMA-2B \cite{zheng2026drivema} & \textbf{8.075} & \textbf{2.662} \\
                \midrule
                Ours [W] & 7.873 & 2.898 \\
                Ours [W] (+BEV task) & \underline{7.902} & \underline{2.891} \\
                \midrule
                Ours [W+N+nSQA] & \underline{7.939} & \underline{2.833} \\
                Ours [W+N+nSQA] (+BEV task) & 7.902 & 2.938 \\
                \bottomrule
            \end{tabular}
            \caption{Performance of various approaches on the \textit{test} set of WOD-E2E \cite{xu2026wod}. ``[W]" and ``[W+N+nSQA]" mean our models are trained on WOD-E2E alone and on WOD-E2E + NAVSIM + nuScenes-QA respectively. Underlined metrics correspond to our model's best configuration given the same training datset.}
            \label{tab:wod-leaderboard}
            \end{table}

        \textbf{On the importance of linguistic data in training.}
            We also provide further evidence that keeping linguistic supervision is important for good performance when adapting pretrained VLMs. For this, we notice that as we add embodiments to the training set through the addition of new public datasets, we increase the degradation of the backbone in its linguistic and semantic reasoning capabilities: models trained on WOD-E2E and on WOD-E2E + NAVSIM, both without the BEV task, on the validation set of nuScenes-QA \cite{qian2024nuscenes} achieve 11.1\% and 7.0\% accuracy respectively, while the base model Qwen 3.5 2B achieves 29.2\%. This could be one factor leading to worse in-dataset performance in terms of FDE when adding NAVSIM to the training data: 5.946 to 6.019 FDE on the validation set of WOD-E2E. Then, after also adding nuScenes-QA (around 10\% of training samples), we recover on FDE (5.851), and multi-dataset training with the BEV task is capable of achieving 8.119 RFS, a value close to 8.13 which is achieved by ground-truth trajectories~\cite{rowe2025poutine}. The improved reasoning quality of the resulting model is verified not only through a nuScenes-QA evaluation of 60.2\% accuracy, which is expected to be high by training on a specific question format, but also through qualitative checks on traffic scene descriptions, available in the Supplementary Material.


            


        \begin{figure}[t]
            \hspace*{-0.7cm}
            \includegraphics[width=1.1\linewidth]{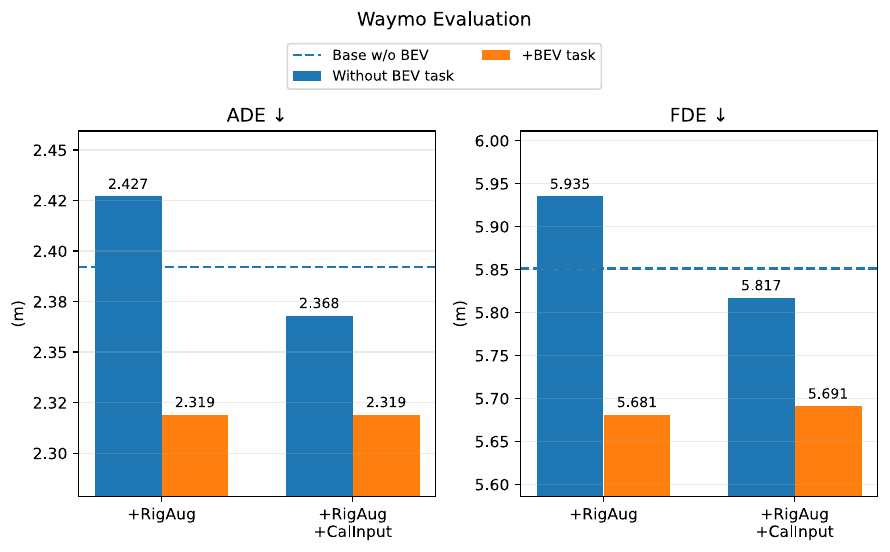}
            \hspace*{-0.7cm}
            \includegraphics[width=1.1\linewidth]{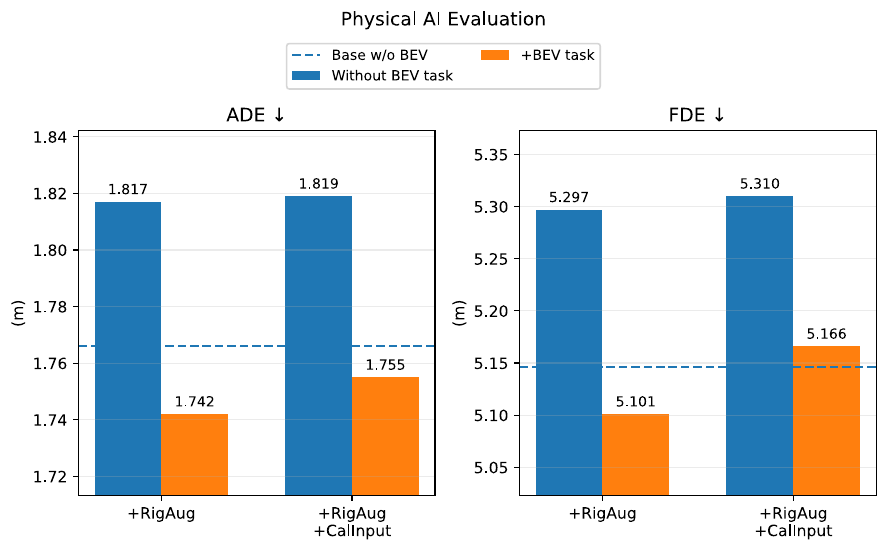}
        
            \caption{Interaction of BEV task with common techniques for dataset robustness, all trainings on WOD-E2E + NAVSIM + nuScenes-QA. As before, runs using the BEV task are able to improve in-dataset WOD-E2E validation results. For zero-shot transfer on Physical AI-1k, common robustness techniques do not lead to improvements unless paired with the BEV task, indicating its usefulness in facilitating learning a unified interface from the increased input signals.}
            \label{fig:bev_ablation_robutness_techs}
        \end{figure}
    

    \subsection{Ablation Studies}
    \label{sec:ablations}

        In this subsection we study what elements are essential for maximal BEV-Forcing effectiveness and how this auxiliary task interacts with other simple robustness techniques.

        \textbf{How does BEV-Forcing interact with common robustness techniques?}
            Since we train on a limited number of datasets, we hypothesize that the model will be able to better learn a unified spatial interface and thus transfer better if together with the BEV task we are able to introduce at least some degree of variability into the camera images that serve as model inputs. To verify this, we study how the BEV task interacts with common ways to introduce robustness to image variations in the model, such as performing image augmentations or injecting calibration information into the model. For the latter, we do it by embedding camera parameters with a simple MLP and adding the resulting vector to its corresponding image patches, multiplied by parameters $\alpha$ that are initialized as zero to minimize the interference with the vision-language alignment of the backbone, as is done in other works \cite{li2026spacedrive}.

            As shown in Fig. \ref{fig:bev_ablation_robutness_techs}, combining the BEV task with such robustness techniques retains improvements in-dataset as measured on the validation set of WOD-E2E. More interestingly, the two techniques mentioned---image augmentations and calibration as input---only lead to better zero-shot transfer when paired with BEV-Forcing. Although improvements compared to the base model without either BEV or other techniques are still modest, this indicates that applying small perturbations on the training procedure that aim at introducing robustness can be more effective when there is an invariant unified spatial interface in which the model can interpret the extra signal in the inputs.

        \textbf{How does the capacity of the BEV head affect performance?}
            Our goal is to instill into the backbone an understanding of the spatial relationship between objects in the scene in a way that affects the planned trajectories. For this, we hypothesize that increasing the capacity of the BEV head, such as by using a full transformer block, leads the target-layer image embeddings to contain sparser spatial information compared to the low capacity version presented in Fig. \ref{fig:bev_head} which consists mostly of just a cross-attention layer with few hidden dimensions compared to the VLM backbone. To test this, we compare two smaller runs: both trained on WOD-E2E and NAVSIM, with our BEV head and a full multi-head attention transformer block. Results are shown in Table \ref{tab:bev-head-capacity}. Notice how the more capable BEV head (precise architecture in Supplementary Material) leads to degraded planning performance in both WOD-E2E which was seen during training and in Physical AI which was not. This gives evidence that forcing the image features themselves to contain denser spatial-related information, sufficient for good BEV construction after simple transformations, is beneficial for planning.

            \begin{table}
            \centering
            \begin{tabular}{l|cc|cc}
                \toprule 
                & \multicolumn{2}{c}{WOD-E2E} &  \multicolumn{2}{c}{PhAI-1k} \\
                \cmidrule(lr){2-3}
                \cmidrule(lr){4-5}
                BEV Head & ADE & FDE & ADE & FDE  \\
                \midrule
                \midrule
                Cross-Attention Only & \textbf{2.376} & \textbf{5.806} & \textbf{1.725} & \textbf{5.073} \\
                Full Transformer Block & 2.441 & 5.959 & 1.805 & 5.308 \\
                \bottomrule
            \end{tabular}
            \caption{Planning performance using the BEV task with BEV heads with different capacities. Notice that the more capable BEV head degrades both in and out of training dataset performance, consistent with our hypothesis that a higher-capacity head places less pressure on the image embeddings themselves to encode spatial information. Metrics not directly comparable to those in Fig. \ref{fig:phai1k-val-zeroshot}, details about this experiment in the Supplementary Material.}
            \label{tab:bev-head-capacity}
            \end{table}

        \textbf{How does BEV-Forcing compare with Spatial Forcing?}
            Finally, to motivate our choice of supervising image features by reconstructing BEV occupancy maps instead of following the procedure in \cite{li2025spatial}, we compare the two methods as before, on both WOD-E2E (in-dataset) and Physical AI (zero-shot) validation sets. After a variety of configurations and hyperparameter tunings, we report in Table \ref{tab:bev-vs-sf} ADE and FDE for the best configurations of each (more details in the Supplementary Material). We find that BEV-Forcing is capable of achieving better metrics in both camera rigs.

            \begin{table}
            \centering
            \begin{tabular}{l|cc|cc}
                \toprule 
                & \multicolumn{2}{c}{WOD-E2E} &  \multicolumn{2}{c}{PhAI-1k} \\
                \cmidrule(lr){2-3}
                \cmidrule(lr){4-5}
                Method & ADE & FDE & ADE & FDE  \\
                \midrule
                \midrule
                BEV-Forcing & \textbf{2.319} & \textbf{5.687} & \textbf{1.783} & \textbf{5.192} \\
                Spatial Forcing \cite{li2025spatial} & 2.341 & 5.766 & 1.818 & 5.264 \\
                \bottomrule
            \end{tabular}
            \caption{Planning performance of BEV-Forcing and Spatial Forcing, training on WOD-E2E + NAVSIM + nuScenes-QA.}
            \label{tab:bev-vs-sf}
            \end{table}






\section{Conclusion}

    In this work we presented the effect of multi-dataset training on driving VLAs as well as BEV-Forcing, a simple technique that generally achieves improved in-dataset planning performance and allows better zero-shot transfer when trained on limited dataset variability, showing it serves as a geometry-aware regularizer. BEV-Forcing supervises the image hidden states at one target layer of the VLM with BEV occupancy maps output by a specialized BEV model, transferring a unified understanding of multi-view cameras to the backbone. We also show how BEV-Forcing improves the effectiveness of common techniques used for robustness training.
    Crucially, with this work we intend to encourage evaluating models and techniques aimed at improving planning with varying amounts of camera rigs or embodiments, including zero-shot transfer evaluations, as an extra way of discerning the applicability or usefulness of the increasing number of techniques proposed each year for improving end-to-end driving.

    \textbf{Limitations and future work.} In this paper we focus on representation learning with image embeddings. Much more can be done, especially with respect to the reasoning trace, action representation, and architectural choices to further improve spatial awareness and geometric transfer. Another direction of future research is exploring how the model behaves when trained with better quality or more informative BEV teacher occupancy maps, such as maps that include drivable area.

{
    \small
    \bibliographystyle{ieeenat_fullname}
    \bibliography{main}
}

\end{document}